\documentclass[11pt]{article}

\usepackage[preprint]{acl}

\usepackage{times}
\usepackage{latexsym}
\usepackage[T1]{fontenc}
\usepackage[utf8]{inputenc}
\usepackage{microtype}
\usepackage{inconsolata}
\usepackage{subcaption}
\usepackage{booktabs}
\usepackage{graphicx}

\usepackage{amsmath}
\usepackage{amssymb}
\usepackage{mathtools}
\usepackage{amsthm}

\usepackage{array}
\usepackage{tabularx}
\newcolumntype{Y}{>{\raggedright\arraybackslash}X}
\usepackage{fvextra}

\usepackage[capitalize,noabbrev]{cleveref}
\usepackage{float}

\theoremstyle{plain}

\theoremstyle{definition}

\theoremstyle{remark}

\title{Towards Training Transformers as a Universal Computer}

\author{
Ruize Xu$^{1}$ \quad
Chenxiao Yang$^{2}$ \quad
Yanhong Li$^{3}$ \quad
David McAllester$^{2}$\\
$^{1}$The University of Chicago \quad
$^{2}$Toyota Technological Institute at Chicago\\
$^{3}$Allen Institute for AI\\
\texttt{richard1xur@uchicago.edu}\\
\texttt{\char`\{chenxiao,mcallester\char`\}@ttic.edu} \quad
\texttt{yanhongl@allenai.org}
}

\begin{document}

\maketitle
\begingroup\renewcommand\thefootnote{}\footnotetext{Published in \emph{Findings of the Association for Computational Linguistics: EMNLP 2026}.}\endgroup

\begin{abstract}

We demonstrate that a small transformer can learn to execute programs in MicroPy, a simplified yet computationally universal programming language. Given procedure definitions together with an expression to evaluate, the transformer predicts small-step execution using PENCIL scaffolding for space-efficient execution within a bounded context window. After training on randomly generated, semantically arbitrary MicroPy programs, the learned transformer generalizes to held-out human-written programs including bit copying and flipping, binary addition and multiplication, and SAT verification and solving. The trained model exhibits out-of-distribution generalization: it correctly evaluates programs drawn from a substantially different distribution than the one used to generate its training data. Since MicroPy can express any computation, our results provide empirical evidence that a standard transformer can be trained to act as a universal computer when paired with PENCIL scaffolding. Project page: \href{https://www.micropy.org}{micropy.org}.

\end{abstract}

\section{Introduction}

Transformer-based language models have achieved remarkable empirical success across a wide range of tasks. Beyond empirical performance, recent theoretical work shows that transformers augmented with Chain of Thought (CoT) \citep{wei2022chain,nye2021show,kojima2022large} are computationally universal: given sufficiently many intermediate steps, a transformer of constant size and depth can simulate Turing machines \citep{merrill2023expresssive,yang2025pencil}. In other words, in principle, any computable problem can be solved by a transformer with CoT.

However, there is a significant gap between what transformers can represent in principle and what they can learn to do through training. In particular, it is unclear whether gradient-based optimization with next-token prediction can produce a transformer that actually exhibits universal computational behavior. This is a non-trivial challenge: even on much narrower tasks, such as length generalization in addition \citep{zhou2023algorithms,zhou2024robust} and constraint-satisfaction problems like Sudoku and 3-SAT \citep{seely2025sudokubench,hazra2024threesat}, current language models with CoT still fall short. The question is also interesting on its own terms: a positive answer would mean there exists a programming language and a task representation under which expressiveness translates into learnability for transformers, and that such a transformer-native combination of language and intermediate-step structure is worth searching for. This motivates a central question: {\bf can we train a transformer to be a universal computer?} Recent work on Neural Computers asks a related question of whether learned systems can acquire computer-like primitives from I/O traces \citep{zhuge2026neuralcomputers}. The closest prior work falls short of answering our question. \citet{npi_reed2016programmer_interpreters} train a neural programmer-interpreter with an LSTM controller, but only for specific, pre-designed functions. \citet{npi_giannou2023looped_transformers} show that a looped transformer can simulate a universal computer (a random access machine), but this is still an expressiveness result.

In this paper, we provide evidence that the answer is yes. We design MicroPy, a simplified yet Turing-complete programming language, and train a small transformer to execute MicroPy programs step by step, using PENCIL scaffolding~\citep{yang2025pencil} for space-efficient execution within a bounded context window. Since MicroPy can express any computation, a model that learns to interpret it acts as a universal computer. In particular:

\paragraph{MicroPy.} MicroPy supports procedure calls, conditionals, recursion, and mutable state over objects with attributes. The input to the model is a set of procedure definitions together with an expression to evaluate; the model then generates, token by token, a step-by-step execution trace. Each step in the trace is either a \emph{retrieval} (looking up a procedure definition, variable binding, or attribute value from the context) or a local \emph{rewrite} (reducing a primitive such as {\tt IF}, {\tt SEQ}, or {\tt LookupAttr} to a simpler expression). A key design principle is that MicroPy's semantics consists of a small, finite set of such retrieval and rewrite rules, so a small subset of training programs is sufficient to cover every rule type at least once.

Beyond language design, a challenge is that CoT's universality comes with an important caveat: intermediate computations accumulate indefinitely, so a task with time complexity $T$ requires context length $\mathcal{O}(T)$ \citep{merrill2023expresssive}. This couples task difficulty directly to context length, making both training and generalization difficult: to handle harder tasks, the model must operate correctly in contexts far longer than those seen during training, i.e. a length generalization problem; moreover, as context grows, the model's ability to retrieve relevant information degrades \citep{liu2024lost}. To address this, we adopt PENCIL scaffolding~\citep{yang2025pencil}.

\paragraph{PENCIL Scaffolding.} PENCIL introduces a reduction rule that reclaims completed intermediate computations from the context, so that the context length is bounded by $\mathcal{O}(S)$, where $S$ is the space complexity of the simulated computation, rather than its time complexity $T$. Specifically, when the context has the form
\begin{equation}
\label{eq1}
\alpha\;\text{[call]}\;\beta\;\rightarrow\;\gamma\;\text{[ret]}
\end{equation}
where [call], $\rightarrow$, and [ret] are special PENCIL tokens, PENCIL replaces the entire sequence with $\alpha\;\gamma$, erasing the intermediate reasoning $\beta$ and keeping only the conclusion $\gamma$. Intuitively, $\alpha$[call]$\beta$ can be viewed as a stack where $\beta$ is the top frame; $\gamma$ is the value returned when that frame is popped. This implements a form of stack discipline analogous to function calls in programming languages: once a sub-computation completes, its intermediate steps are discarded and only the result is kept. In practice, PENCIL enables the model to execute much longer programs within a fixed context window. \Cref{fig:pencil-space-efficiency} gives a space-efficiency analysis on the held-out tasks, comparing PENCIL against append-only traces under the same 4096-token budget.

\paragraph{Experiments.} We trained a 59.5M-parameter decoder-only transformer with RoPE~\citep{su2024roformer} on execution traces generated from random MicroPy programs, capped at 128 trace lines per program. We evaluated on held-out, human-written MicroPy programs that are substantially longer and semantically meaningful: bit copying and flipping, binary addition and multiplication (with traces up to 7,552 lines), and SAT verification and solving. The model achieves 100\% token-level accuracy on all tasks within the tested context window. The near-zero overlap between the training-program distribution and the held-out evaluation suite (\cref{fig:program-length-hist}) provides direct evidence of both compositional and length generalization.

To summarize, our main contributions are:
\begin{enumerate}\setlength{\itemsep}{2pt}\setlength{\parskip}{0pt}\setlength{\topsep}{2pt}
\item We design MicroPy, a computationally universal language.
\item We provide the first empirical evidence that a standard transformer can be trained to simulate a universal computer by running a Turing-complete language under PENCIL scaffolding, up to the model's context window.
\item We demonstrate strong generalization: 100\% token-level accuracy on held-out human-written programs with traces up to 60$\times$ longer than those seen during training, under PENCIL scaffolding.
\end{enumerate}

\paragraph{Broader Implication.} It is well known that generalization requires inductive bias (the no free lunch theorem; \citealt{Wolpert_1996}), and not all universal priors are equally efficient. Our results suggest that transformers with suitable scaffolding like PENCIL may possess an inductive bias well-suited to phrase structure and compositional semantics, the core structures shared by all high-level programming languages, each of which is computationally universal. Since natural language exhibits analogous compositional structure, this may shed light on why transformers, with context management scaffoldings, are so effective at natural language tasks: the same implicit bias that enables learning a programming language interpreter may also facilitate learning the compositional semantics of natural language.

\begin{figure*}[t!]
  \centering
  \includegraphics[width=\textwidth]{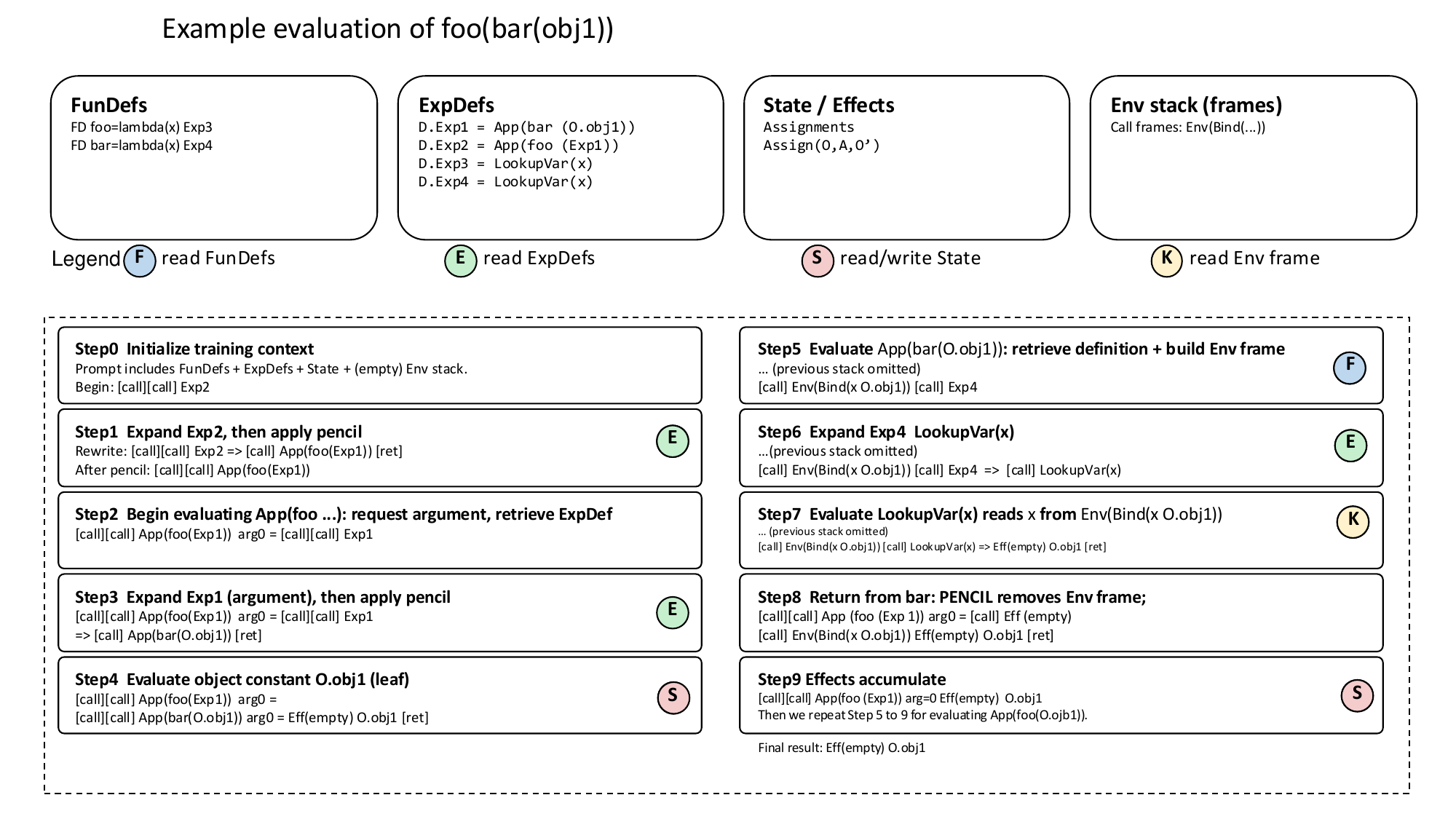}
  \caption{Evaluation of a MicroPy expression.  We define foo and bar to both be the identity function and evaluate foo(bar(O)) where O is an object symbol. The general rules for evaluation are described in the text. More details about what each step is doing and the exact definitions on this table can be found in \cref{app:example-data}.}
  \label{fig:overview-figures}
\end{figure*}

\section{MicroPy and Program Execution}

In this section, we describe the MicroPy language and how programs are executed. We first present the two execution primitives, retrieval and rewriting (\cref{sec:retrieval-rewriting}), then describe the evaluation of procedure calls and tail recursion (\cref{sec:procedure-calls}), and finally give the formal definition of the MicroPy language (\cref{sec:micropy-language}).

\subsection{Retrieval and Rewriting}
\label{sec:retrieval-rewriting}

The evaluator relies on two conceptual primitives: retrieval and rewriting.

We first focus on retrieval.  Our evaluator uses
five kinds of symbols: function symbols, expression symbols, variable symbols,
object symbols and attribute symbols. The context contains definitions for function symbols, expression symbols,
and variable symbols. For example a function definition is (some version of) $f(x_1,\ldots,x_n) = e$ where $f$ is a function symbol,
each $x_i$ is a variable symbol, and $e$ is an expression symbol.  An expression definition has a form such as $e_0 = \mathrm{IF}(e_1,e_2,e_3)$ where $e_0$, $e_1$, $e_2$ and $e_3$ are expression symbols.  Each expression definition specifies a primitive of the language and expression symbol arguments for that primitive.  The primitives include looking up the value of a variable.
Variable values are held in stack frames.  The value of a variable must be retrieved from the most recent stack frame.
The context also contains a state consisting of triples $\langle O_1,A,O_2 \rangle$ stating that the value of attribute $A$ of object $O_1$ is the object $O_2$.  The language supports effects which can change the value of an attribute. As with variables, we need the most recent
assignment of a value to a given attribute of a given object.
The transformer must be capable of retrieving the values of function symbols, expression symbols, variable symbols and attribute values.  Each retrieval does a replacement.  For example $\mathrm{LookupVar}(x)$ at the end of the context gets replaced by
the value of $x$ in the current stack frame (using a PENCIL rewrite).

In addition to the required retrievals, we need to be able to compute within an active stack frame. This involves computing the value of expressions under the variable values in the stack frame.\footnote{Variable symbols are analogous to registers
and a definition of an expression symbol is analogous to a machine instruction in a RAM model of computation. Here there are no primitive arithmetic operations. However, arithmetic can be implemented as computation on bit strings.}  We can think of computation within a given
stack frame as a rewriting process.
For example $\mathrm{IF}(e_1,e_2,e_3)$
gets rewritten to $e_2$ if the value of $e_1$ is the special object $\mathrm{true}$ and rewritten to $e_3$
if the value of $e_1$ is not $\mathrm{true}$.

\subsection{Procedure Calls and Tail Recursion}
\label{sec:procedure-calls}

\cref{fig:overview-figures} shows an example of evaluating an expression using the
transformer context for the required memory.  Here we outline the general steps of evaluation.
A call to a procedure is a context of the form
\begin{equation}
\label{tail1}
\alpha\,\mathrm{[call]}\,e_1\,\mathrm{[call]}\,\mathrm{Apply}(f,E_1,\ldots,E_n)
\end{equation}
where $e_1$ is a ``state'' specifying the values of attributes of objects,
$f$ is a function symbol and $E_1,\ldots,E_n$ are expression symbols.
This is eventually rewritten to
\begin{equation}
\label{tail2}
\alpha\,\mathrm{[call]}\,e_2\,\mathrm{[call]}\,\mathrm{TailApply}(f,O_1,\ldots,O_n)
\end{equation}
where $O_1,\ldots,O_n$ are object symbols and $e_2$ is a new attribute-value state
that combines $e_1$ with effects performed during the evaluation of the argument expressions.

In the normal case (where there is no tail recursion optimization), and for $f$ defined by
$f(x_1,\ldots,x_n) = E$, we have that \eqref{tail2} gets rewritten to
\begin{equation}
\label{tail3}
\resizebox{0.80\columnwidth}{!}{$\alpha\,\mathrm{[call]}\,e_2\,\mathrm{[call]}\,\mathrm{BIND}((x_1,O_1),\ldots,(x_n,O_n))\,\mathrm{[call]}\,E$}
\end{equation}
Here $\mathrm{BIND}((x_1,O_1),\ldots,(x_n,O_n))$ is a stack frame holding the values for the ``local variables''
$x_1,\ldots,x_n$.
It is possible that \eqref{tail3} gets rewritten to
\begin{equation}
\label{tail4-frame}
\resizebox{0.80\columnwidth}{!}{$\alpha\,\mathrm{[call]}\,e_2\,\mathrm{[call]}\,\mathrm{BIND}((x_1,O_1),\ldots,(x_n,O_n))\,e_3\,O$}
\end{equation}
where $O$ is an object symbol. In this case \eqref{tail2} gets rewritten to
\begin{equation}
\label{tail4-return}
\alpha\,e_4\,O
\end{equation}
where $e_4$ is a combination of the effects $e_2$ and $e_3$.

However, an alternative possibility is that \eqref{tail2} gets rewritten to
\begin{equation}
\label{tail5}
\resizebox{0.80\columnwidth}{!}{$\alpha\,\mathrm{[call]}\,e_2\,\mathrm{[call]}\,\mathrm{BIND}((x_1,O_1),\ldots,(x_n,O_n))\,e_3\,c$}
\end{equation}
where $c = \mathrm{TailApply}(g,O_1,\ldots,O_m)$.  In this case \eqref{tail2}
gets rewritten to
\begin{equation}
\label{tail6}
\alpha\,\mathrm{[call]}\,e_4\,\mathrm{[call]}\,\mathrm{TailApply}(g,O_1,\ldots,O_m)
\end{equation}
Note that \eqref{tail6} has the same form as \eqref{tail2}.  If $g=f$ we get a form of iteration.

\subsection{The MicroPy Language}
\label{sec:micropy-language}

As described above, MicroPy is a Turing-complete language whose state is built from symbols; it can be used as an object-oriented programming language by treating these symbols as objects with attributes. A state of the system is represented by a sequence of triples $\langle O_1,A,O_2\rangle$
stating the value of the attribute $A$ of object $O_1$ is the object $O_2$.
A MicroPy procedure definition has the form
\[
f(x_1,\ldots,x_n) = E
\]
where $x_1,\ldots,x_n$ are variable symbols and $E$ is a MicroPy expression symbol.  An expression definition has one of the forms
$E_0 = \mathrm{App}(f,E_1,\ldots,E_n)$ or $\mathrm{LookupVar}(x)$, or
$E_0 = \mathrm{op}(E_1,\ldots,E_n)$ where $\mathrm{op}$ is a primitive operation and each $E_i$ is either an expression symbol or an attribute symbol.  The primitive operations are the following.

\begin{itemize}\setlength{\itemsep}{2pt}\setlength{\parskip}{0pt}
\item {\tt IF}$(E_1,E_2,E_3)$: conditional expression.
\item {\tt EQUAL}$(E_1,E_2)$: equality test.
\item {\tt HasAttr}$(E,A)$: returns {\tt true} if the value of $E$ has a value assigned for attribute $A$.
\item {\tt Assert}$(E_1,A,E_2)$: updates the value of attribute $A$ of the value of $E_1$ to the value of $E_2$.
\item {\tt LookupAttr}$(E,A)$: returns the value of attribute $A$ of the value of $E$.
\item {\tt SEQ}$(E_1,E_2)$: evaluates $E_1$ for effect, then evaluates and returns $E_2$.
\item {\tt Try}$(E_1,E_2)$: evaluates $E_1$; if the result is the special object $\mathrm{fail}$, all effects of $E_1$ are rolled back and $E_2$ is evaluated instead. Otherwise, the result and effects of $E_1$ are kept.
\end{itemize}

In all test data the state is initialized to contain linked lists defined by the attribute "next".  For example a list of "bit objects" would be represented by a state with attribute values such as {\tt Assign(BIT5,NEXT,BIT6)}. Bit strings can be
represented by assigning a value to each bit in a list of bit objects.  For example {\tt Assign(BIT7,VALUE,TRUE)}. Bit strings can be interpreted as binary numbers and arithmetic operations such as addition and multiplication can be computed on bit strings.

It is also possible to initialize the state to a long linked list of objects representing a heap memory.  We can have an object "freepit"
whose value is one of the objects in the heap.  This allows one to implement heap allocation of data structures.  We can then implement
list operations such as appending two lists.

Although not done in our current experiments, it is possible to replace the assert primitive with a pairing primitive {\tt Pair}$(e_1,e_2)$.
The pairing primitive would create a new object with "first" and "second" attributes that are assigned the values of $e_1$ and $e_2$ respectively.
With a pairing primitive one could write purely functional (effect-free) programs.
We see no reason to believe that our experimental results would be any different in a purely functional language.

\paragraph{Model input and prediction target.}
\Cref{fig:overview-figures} breaks the evaluation of \texttt{foo(bar(obj1))} into local transitions. For the first transition in that example, the model receives the following flat token sequence containing the procedure dictionary (\texttt{FunDefs}), expression dictionary (\texttt{ExpDefs}), accumulated pending state updates (\texttt{Eff}), active variable-binding frames (\texttt{Env}), and execution stack. Tokens through \texttt{[EOP]} form the input, and the continuation after \texttt{[EOP]} is the prediction target (line breaks are added for readability):
\begin{Verbatim}[fontsize=\scriptsize]
MODEL INPUT:
FunDefs (
  FD foo = lambda ( x ) Exp3 ;
  FD bar = lambda ( x ) Exp4 ; )
********* ExpDefs (
  D . Exp1 = App ( bar ( O . obj1 ) )
  D . Exp2 = App ( foo ( Exp1 ) )
  D . Exp3 = LookupVar ( x )
  D . Exp4 = LookupVar ( x ) )
****** Eff ( empty )
********* Env ( ) [call] [call] Exp2 [EOP]
PREDICTION TARGET:
=> [call] App ( foo ( Exp1 ) ) [ret]
\end{Verbatim}
PENCIL applies the predicted transition to construct the next model input, and evaluation proceeds one local transition at a time. The complete walkthrough appears in \cref{app:example-data}.

\newcommand{\Effects}{\mathcal{U}}

\DefineVerbatimEnvironment{MicropyBlock}{Verbatim}{
  fontsize=\scriptsize,
  breaklines=true,
  breakanywhere=true,
  frame=single,
  framesep=2mm
}
\section{Training and Experiments}

This section describes how we generate training data (\cref{sec:data-generation}), the model and training setup (\cref{sec:training-setup}), and the experimental results (\cref{sec:results}).

\subsection{Data Generation}
\label{sec:data-generation}

Every PENCIL step in a MicroPy execution (\cref{eq1}) is one of two things: a \emph{retrieval} that reads a definition or binding from one of the context blocks (\texttt{FunDefs}, \texttt{ExpDefs}, \texttt{Env}, \texttt{Eff/State}), or a \emph{rewrite} that reduces a primitive (\texttt{If}, \texttt{Seq}, \texttt{LookupAttr}, and so on) using values and effects already on the stack (\cref{fig:overview-figures}). There are finitely many of each, so a model that learns this fixed set of moves can execute any MicroPy program up to the context window. Training data should therefore cover these moves directly, not particular programs.

\paragraph{Online trace generation.}
Training data is generated online: at each step we sample a MicroPy expression, run it, and turn one of its PENCIL trace lines into a (prompt, completion) pair. We call each such pair, which implements one local retrieval or rewrite before PENCIL reduction, a \emph{trace step}. We use two samplers. The first samples whole MicroPy programs directly. The second samples a \emph{plan}, a target stack configuration, then compiles it into an expression that hits that configuration and uses only the matching trace line.

\paragraph{Program sampler.}
The first sampler $\mathcal{G}_{\mathrm{prog}}$ is a probabilistic context-free grammar (PCFG) over MicroPy procedure definitions and expressions, capped by global depth/size/effect/binding budgets. We sample a program $e \sim \mathcal{G}_{\mathrm{prog}}$, execute it in the MicroPy interpreter, and turn each PENCIL rewrite in the trace into a training example. This gives wide syntactic coverage but no direct handle on rare runtime configurations.

\paragraph{Plan sampler.}
The second sampler targets a specific shape directly. We first sample a \emph{plan} $\pi$ that says what one trace step should look like at the deepest call-stack tail. A plan consists of a stack skeleton together with binding and effect requirements that the realized program must satisfy. The skeleton is an ordered sequence of frame labels printed after \texttt{[call]} along the deepest call-stack tail; it ends in one leaf retrieval (\textsc{LookupVar}, \textsc{LookupAttr}, or \textsc{TailApp}) wrapped by some number of continuation frames (\textsc{Seq}, \textsc{If}, \textsc{Try}, \textsc{Env}, or \textsc{Eff}). The plan sampler $\mathcal{G}_{\mathrm{plan}}$ draws the skeleton recursively: start from a leaf and, with each step, either stop or wrap the current skeleton with one more continuation frame, until a depth budget is reached. The binding and effect requirements are then turned into concrete \texttt{Env(\ldots)} bindings and \texttt{Eff(\ldots)} / \texttt{State/Effects} assertions during compilation. When the skeleton is compiled into an expression, we route the recursive descent through the first child of \textsc{Seq}, the condition of \textsc{If}, or the second child of \textsc{Try}, so that the sampled labels land contiguously in the deepest tail at runtime. \cref{fig:plan-trace-example} shows an example plan with a matched trace step.

\paragraph{From plan to training example.}
Compiling a plan produces an executable expression $e_{\pi}$. Running $e_{\pi}$ generates a trace whose deepest call-stack tail matches the planned skeleton at exactly one step, and we use that step's prompt and completion as the training example; other steps from the same execution are discarded. To vary what surrounds the planned step (the outer stack prefix, unrelated bindings, distractor effects), we optionally drop $e_{\pi}$ into a larger expression sampled by $\mathcal{G}_{\mathrm{ctx}}$ and execute that instead. The remaining training budget is filled with examples from $\mathcal{G}_{\mathrm{prog}}$.

\begin{figure*}[t!]
  \centering
  \includegraphics[width=\textwidth]{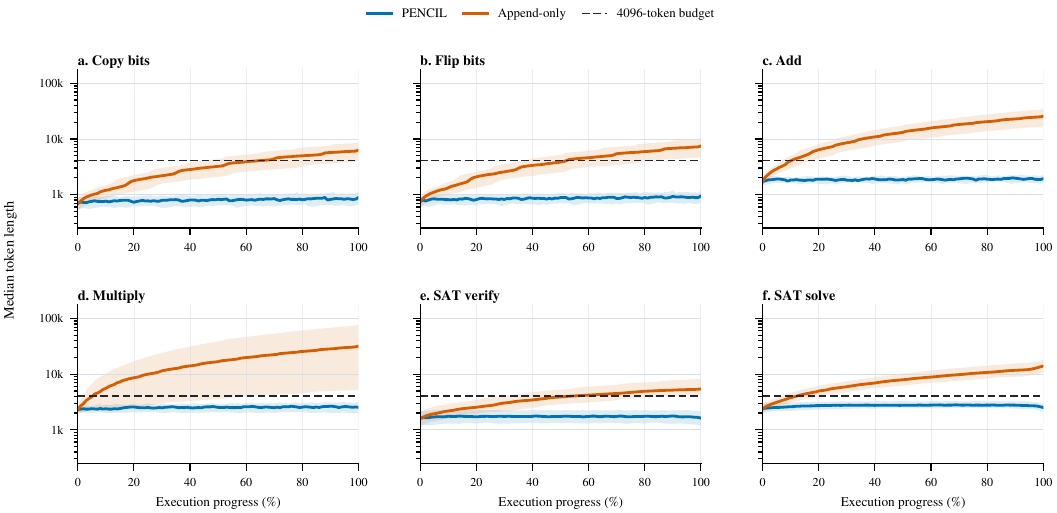}
  \caption{
  PENCIL bounds the live token length during long MicroPy executions.
  For each held-out human-written task, we compare the median token length over normalized execution progress under PENCIL scaffolding and under an append-only transcript that retains all previous completions.
  Shaded regions show the interquartile range across programs, and the dashed line marks the 4096-token budget used by the model.
  Without PENCIL, many simple programs require very long live contexts: append-only traces grow with execution progress and can exceed the 4096-token budget before completion, substantially increasing the difficulty of training on longer executions.
  This analysis isolates the space-efficiency role of PENCIL: both curves use the same evaluated programs and next-step completions, differing only in prior-text retention.
  }
  \label{fig:pencil-space-efficiency}
\end{figure*}

\begin{figure}[t]
  \centering
  \begin{minipage}{0.95\linewidth}
    \begin{MicropyBlock}
PLAN :
  Eff -> Env -> Seq -> Eff -> Env -> Eff -> Eff -> If -> Try -> Seq -> Eff -> TailApp

MATCHED STEP (general stack):
********* Env ( )
[call] Eff ( Assertion ( O . env0 Att . value False ) Assertion ( O . env1 Att . value False )
            Assertion ( O . noise0 Att . k O . env1 ) Assertion ( O . noise1 Att . k O . env1 ) )
[call] Env ( Bind ( p O . plan0 ) Bind ( b O . ljO5H0 ) Bind ( x0 O . env0 ) Bind ( x1 O . env1 ) ... )
[call] Seq ( Exp22 , False ) x =
[call] Eff ( empty )
[call] Env ( ... Bind ( y0 O . env0 ) Bind ( y1 O . env2 ) )
[call] Eff ( Assertion ( O . env0 Att . value True ) )
[call] Eff ( Assertion ( O . env2 Att . value False ) )
[call] If ( Exp38 False False ) test =
[call] [call] Try ( Exp34 Exp37 ) second =
[call] [call] Seq ( Exp36 , Exp32 ) x =
[call] Eff ( empty )
[call] TailApp ( _tailapp_id Args ( O . env0 ) )

result:
=> [call] let ( z ) be ( O . env0 ) in Exp10 [ret]
    \end{MicropyBlock}
  \end{minipage}
  \caption{Sampled plan and matched trace step (general stack). Each \texttt{[call]} line is followed by the frame-label sequence specified by the plan. \texttt{Env} frames contain active variable bindings, and \texttt{Eff} frames contain accumulated pending state updates. Some binding lists are omitted for readability.}
  \label{fig:plan-trace-example}
\end{figure}

\paragraph{Program sampler parameterization.}
The program-level sampler $\mathcal{G}_{\mathrm{prog}}$ is implemented via two coupled samplers:
(i) a \emph{surrounding-code sampler} $\mathcal{G}_{\mathrm{ctx}}$ that generates enclosing expression templates with a designated insertion point, and
(ii) an \emph{expression sampler} $\mathcal{G}_{\mathrm{expr}}$ that generates the side expressions used by those templates.
Rule probabilities are renormalized under depth/size/effect/binding budgets.

\paragraph{Surrounding-code sampler (general stack section).}
Templates are drawn from a separate sampler $\mathcal{G}_{\mathrm{ctx}}$: starting from a hole $\Box$ where the planned expression will be inserted, $\mathcal{G}_{\mathrm{ctx}}$ recursively wraps the hole with a compositional MicroPy operator (\textsc{Seq}, \textsc{If}, \textsc{Try}, \textsc{App}, \textsc{Let}, \textsc{LookupAttr}), filling the non-hole argument slots with side expressions drawn from $\mathcal{G}_{\mathrm{expr}}$. \textsc{If}/\textsc{Try} place the hole on \emph{both} branches so the planned expression is reached under either sampled branch outcome. Wrapping continues under depth/size budgets until the hole becomes a complete enclosing context, at which point the planned expression is substituted in. As above, we retain only executions whose realized trace contains the planned configuration in the intended stack position.

\paragraph{Expression sampler (side expressions).}
Side expressions are drawn from a sampler $\mathcal{G}_{\mathrm{expr}}$ defined recursively: given the planned bindings and effect requirements, at each step $\mathcal{G}_{\mathrm{expr}}$ either places a compositional MicroPy operator (\textsc{Seq}, \textsc{If}, \textsc{Try}, \textsc{App}, \textsc{Let}, \textsc{LookupAttr}) and recurses into its argument slots, or stops at a leaf that reads a planned binding (\textsc{LookupVar}) or emits a Bool/object constant. Probabilities over the operator and leaf categories give us per-op control over which MicroPy primitives the sampled expression exercises.

\label{sec:micropy-experiments}

\begin{table*}[t]
\centering
\scriptsize
\setlength{\tabcolsep}{3pt}
\resizebox{\textwidth}{!}{%
\begin{tabular}{lrrrrrrrrr}
\toprule
Task & bit-len & ctx$_{\min}$ & ctx$_{\max}$ & lines$_{\min}$ & lines$_{\max}$ & \#programs & \#trace lines & avg lines/program & accuracy \\
\midrule
copy\_bits & 2--10 & 215  & 725  & 78  & 422   & 9 & 2,250  & 250.0 & 100\% \\
flip\_bits & 2--10 & 253  & 774  & 108 & 572   & 9 & 3,060  & 340.0 & 100\% \\
RPC\_add   & 2--10 & 722  & 1514 & 469 & 2365  & 9 & 12,753 & 1,417.0 & 100\% \\
RPC\_mult  & 2--10 & 971  & 2197 & 80  & 7552  & 9 & 24,379 & 2,708.8 & 100\% \\
\midrule
\multicolumn{6}{r}{Total (bit-length tasks)} & 36 & 42,442 & 1,178.9 & 100\% \\
\bottomrule
\end{tabular}
}
\caption{Test dataset statistics for bit-length tasks. Context length is tokens per step; lines$_{\min/\max}$ are per-program execution lengths; \#trace lines is the total number of step-wise prompt/completion pairs. Accuracy is the token-level success rate.}
\label{tab:bitlen-stats}
\end{table*}

\begin{table*}[t]
\centering
\scriptsize
\setlength{\tabcolsep}{3pt}
\resizebox{\textwidth}{!}{%
\begin{tabular}{lrrrrrrrrrr}
\toprule
Task & \#vars & \#clauses & ctx$_{\min}$ & ctx$_{\max}$ & lines$_{\min}$ & lines$_{\max}$ & \#programs & \#trace lines & avg lines/program & accuracy \\
\midrule
sat\_solve  & 2--4 & 1--6 & 941  & 2091 & 188 & 5020 & 180 & 172,251 & 957.0 & 100\% \\
sat\_verify & 2--4 & 1--6 & 476  & 1398 & 69  & 628  & 36  & 10,015  & 278.2 & 100\% \\
\midrule
\multicolumn{7}{r}{Total (SAT tasks)} & 216 & 182,266 & 843.8 & 100\% \\
\bottomrule
\end{tabular}
}
\caption{Test dataset statistics for SAT tasks. Context length is tokens per step; lines$_{\min/\max}$ are per-program execution lengths; \#trace lines is the total number of step-wise prompt/completion pairs. Accuracy is the token-level accuracy.}
\label{tab:sat-stats}
\end{table*}

\subsection{Model and Training Setup}
\label{sec:training-setup}

\paragraph{Vocabulary and tokenizer.}
We use a whitespace tokenizer with a fixed vocabulary containing \textsc{MicroPy}-specific syntax and formatting tokens (e.g.\ \texttt{``If "} and the delimiter \texttt{``[call] "}). A fixed vocabulary is appropriate here because the language emits a small, finite set of syntactic forms; we trade flexibility for unambiguous token boundaries during step-wise prediction. The implications of this choice are discussed in the Limitations section.

\paragraph{Model and optimization.}
We train a 59.5M-parameter decoder-only transformer (depth $10$) with rotary position embeddings \citep{su2024roformer} on traces from the two samplers of \cref{sec:data-generation}. Following the NanoChat recipe \citep{karpathy2025nanochat}, we use AdamW for token embedding and unembedding \citep{loshchilov2018decoupled} and the Muon optimizer for transformer-block matrices \citep{jordan2024muon}, with learning rates $0.2/0.004$ (Adam) and $0.02$ (Muon). We train for $250{,}000$ iterations at maximum sequence length $4096$ and global batch size $1{,}966{,}080$ tokens, with $5\%$ linear warmup, final-LR decay to $5\%$ of peak, and no weight decay. All reported numbers use greedy next-token decoding under a single training seed; hyperparameters were not selected by sweep but follow the NanoChat defaults. Training runs on a single 8-GPU node of a university cluster in bfloat16 (about one node-week of wall-clock time). The training-data generator, trained checkpoint, evaluation programs, and program-execution code will be released under a permissive open-source license through the project page upon publication. Full compute, data-volume, and dataset-statistics details are reported in \cref{app:repro}.

\paragraph{Evaluation.}
We evaluate on actual human-written MicroPy functions: \textsc{Copy}, \textsc{Flip}, \textsc{Addition}, \textsc{Multiplication}, \textsc{SAT-VERIFY}, and \textsc{SAT-SOLVER}. The function definitions are in \cref{app:micropy-functions} and per-task statistics in \cref{tab:bitlen-stats} and \cref{tab:sat-stats}. We achieve 100\% token-level accuracy on each task within the tested context lengths.

\begin{figure*}[!t]
  \centering
  \begin{subfigure}[t]{0.55\textwidth}
    \centering
    \includegraphics[width=\textwidth]{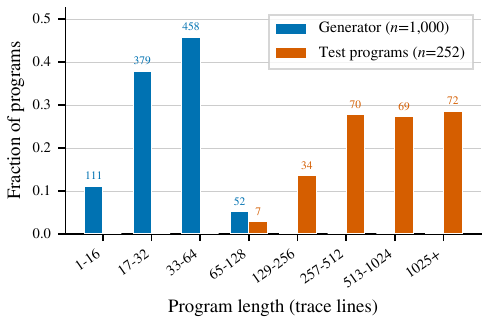}
    \caption{Program length distribution for generator-sampled training programs and the held-out evaluation suite ($1{,}000$ sampled vs.\ $252$ human-written programs). The near-zero overlap beyond the $65$--$128$ bin shows that the model is trained on short programs but evaluated on traces that are often much longer.}
    \label{fig:program-length-hist}
  \end{subfigure}\hfill
  \begin{subfigure}[t]{0.40\textwidth}
    \centering
    \includegraphics[width=0.85\textwidth]{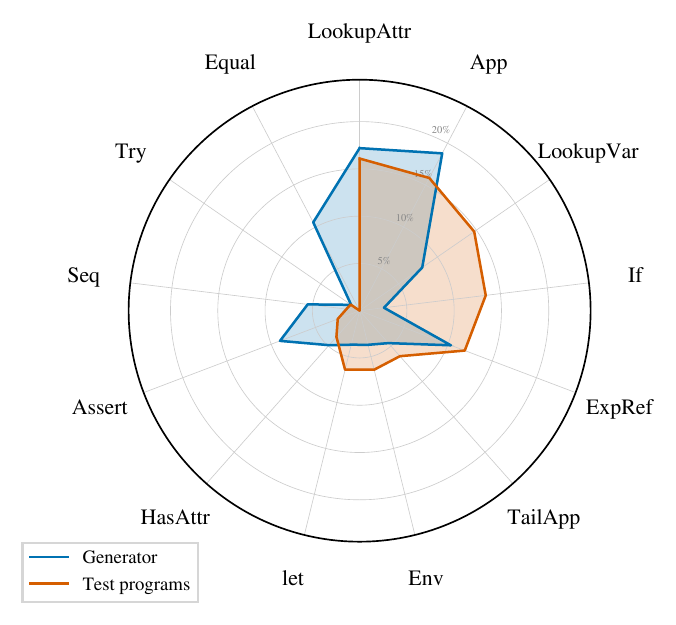}
    \caption{Pooled operation-profile comparison: training generator (blue) vs.\ held-out suite (orange). The generator covers every operation type; the held-out programs place more mass on \texttt{If}, \texttt{LookupVar}, \texttt{ExpRef}, and \texttt{TailApp}.}
    \label{fig:pooled-last-op-radar}
  \end{subfigure}
  \caption{Experiment summary plots: (a) length gap between training and held-out programs; (b) operation-profile gap.}
  \label{fig:experiment-summary-plots}
\end{figure*}

\subsection{Results}
\label{sec:results}

\paragraph{Compositional and length generalization.}
\Cref{fig:program-length-hist} compares the program-length distribution of generator-sampled training programs with the held-out evaluation suite. In theory we do not have to train on any concrete program. In practice, we cap generator-sampled training programs at 128 trace lines. The held-out evaluation suite (\cref{tab:bitlen-stats} and \cref{tab:sat-stats}) contains much longer total executions, with per-program traces reaching 7{,}552 lines (about $60\times$ longer than the training cap on total trace length). PENCIL keeps the per-step input bounded at $4096$ tokens regardless of total trace length, so the $60\times$ figure refers to trace length, not to the model's per-step input. The near-zero overlap between the training-generator distribution and the test suite provides a direct visualization of length generalization at the trace level.

\paragraph{Coverage of MicroPy primitives.}
Our plan-based generator is designed to cover all procedure combinations: for any retrieval/rewrite pattern permitted by MicroPy's small-step semantics, specified as a target call-stack skeleton with concrete Env/Eff binding shapes and retrieval channels, there exists a plan whose execution realizes that runtime configuration. The pooled operation-profile comparison and per-task breakdowns in \cref{fig:per-task-radar} show that the generator covers every operation type used by the test tasks. Without a budget and subject only to the length constraint, this implies a non-zero sampling probability for any finite MicroPy program. The remaining gap to larger programs is therefore an engineering constraint of the context window (see Limitations).

\section{Related Work}

Our work connects to several lines of research. \textbf{CoT and bounded context:} Transformers with CoT are theoretically universal but the context grows linearly with computation time \citep{merrill2023expresssive,li2024chain}; PENCIL reclaims intermediate steps to bound context by space rather than time \citep{yang2025pencil}. \textbf{Length generalization:} Prior work improves extrapolation via task-specific representations such as position coupling \citep{cho2024positioncoupling,mcleish2024rightembeddings} or Turing-machine-style decomposition \citep{hou2024turingprograms,abbe2024globality}; MicroPy applies the decomposition principle at the level of a programming language, training one model for all tasks. \textbf{Neural interpreters and computers:} NPI \citep{npi_reed2016programmer_interpreters,npi_cai2017recursion} and learned execution \citep{npi_zaremba2014learning_execute,nye2021show,npi_liu2023codeexecutor} explore learned program execution but rely on specialized architectures or task-specific training; we use a standard transformer with next-token prediction. \citet{graumoya2024learning} meta-train sequence models on outputs of random BrainPhoque universal-machine programs to approximate universal next-symbol prediction. Their predictor observes emitted symbols but not the generating program or internal machine configuration; our model instead receives explicit MicroPy definitions and execution state and is supervised on the next local interpreter transition. \citet{zhuge2026neuralcomputers} propose Neural Computers that learn interface and control primitives from I/O traces in CLI and GUI settings; in contrast, we train a transformer to execute a formal, Turing-complete programming language with explicit operational semantics. \textbf{Compositional generalization:} Classical benchmarks such as SCAN, CFQ, and COGS test novel compositions of familiar primitives \citep{lake2018generalization,keysers2020measuring,kim2020cogs}; in our setting the composition rule is an operational semantics \citep{drozdov2022compositional,dziri2023faith}. \textbf{External memory:} Retrieval augmentation and long-term memory modules \citep{lewis2020rag,wang2024augmenting} address bounded context by storing information externally; our focus is on which execution state must stay in context for the next step to be well-defined. See \cref{app:related-work} for a detailed discussion.

\section{Conclusion}

We showed that a standard decoder-only transformer, trained with next-token prediction on randomly generated MicroPy traces, can learn to act as a neural interpreter that generalizes to held-out human-written programs with 100\% token-level accuracy. These results provide empirical evidence that, within a finite context window, a Transformer with PENCIL can learn the operational semantics of a computationally universal language. We view computational semantics as a useful lens for language-model agents and long-horizon-task systems, which similarly rely on transformers to maintain structured state and execute step by step over thousands of tokens. Limitations include restriction to the MicroPy language and a fixed context window; extending to richer languages, longer contexts, and reasoning about program behavior without execution are promising directions for future work.

\section*{Limitations}
\label{sec:finite-vocab}

\paragraph{Finite execution regime.}
Our experiments use a fixed 4096-token context window and evaluate a finite suite of held-out programs. PENCIL reduces live context from computation time to computation space, but programs whose live state exceeds the context window remain outside the evaluated regime. The reported results are empirical and do not establish arbitrary-length correctness.

\paragraph{Training and representation.}
The model is trained on synthetic traces from one procedural generator under a single training seed. Its behavior on execution shapes not represented by that generator remains unmeasured; \cref{app:failure-case-tail-eff} gives examples from ablated training distributions. The model also assumes the fixed prompt and trace format used in training. Its whitespace tokenizer has a finite vocabulary, including fixed pools of object and attribute symbols, which bounds the number of distinct names in one serialized program. We leave alternative tokenizations and broader training distributions to future work.

\paragraph{Language scope.}
MicroPy is a small language designed to expose retrieval, rewriting, mutable state, and stack management in a controlled setting. Our experiments do not test transfer to production languages or compare alternative computationally universal languages. Extending the study to established languages and additional execution interfaces is future work.

\section*{Ethical Considerations}

\paragraph{Risks.}
Our contribution is a controlled study of program execution by a small transformer trained on synthetic traces of a designed language. The trained model emits step-wise execution rewrites for MicroPy programs only and has no capability to access external systems, generate natural language, or execute code in production languages such as Python or C. We are not aware of dual-use risks specific to this artifact beyond those general to small language models trained on synthetic data. Insofar as our results support the broader claim that transformers can learn compositional, universal computation under suitable scaffolding, the same observation also applies to more capable systems, where the standard concerns about misuse of learned program executors (e.g., automated reasoning over untrusted inputs) apply.

\paragraph{Broader impact and intended use.}
The MicroPy interpreter is intended as a research vehicle for studying compositional and length generalization in transformer language models. We do not advocate using it as a substitute for verified interpreters in safety-critical settings. The released artifacts (training-data generator, model checkpoints, and evaluation suite) are intended for academic reproduction and follow-up research, and we encourage downstream users to preserve this scope.

\paragraph{Data and artifacts.}
All training data is synthetically generated by a procedural sampler we describe in \cref{sec:data-generation} and contains no human-subject data, no personally identifying information, and no scraped web text. The held-out evaluation programs (\textsc{Copy}, \textsc{Flip}, \textsc{Addition}, \textsc{Multiplication}, \textsc{SAT-Verify}, \textsc{SAT-Solver}) are author-written MicroPy programs. We use the NanoChat training stack of \citet{karpathy2025nanochat} (MIT-licensed open source), the Muon optimizer of \citet{jordan2024muon}, and the PENCIL scaffolding rule of \citet{yang2025pencil}; all are used consistent with their published intended use. Our generated dataset, our trained checkpoint, and the evaluation programs will be released under a permissive license upon publication.

\section*{Acknowledgments}

We thank Tianchong Jiang, Anxin Guo, Xiaodan Du, and Zhiyuan Li for helpful discussions, and the staff at TTIC for maintaining the compute cluster used in our experiments. We also thank the anonymous reviewers for their constructive feedback.

\bibliography{ref,ref_neural_programmer_interpreter}

\clearpage
\appendix

\section{Reproducibility, Compute, and Data Statistics}
\label{app:repro}

\paragraph{Compute.}
Training was run on a single 8-GPU node of an internal university cluster, using bfloat16 autocast and distributed data parallelism via \texttt{torchrun}. The cluster contains a mixture of NVIDIA RTX~6000 Ada, L40S, and H100 GPUs; the reported training run used whichever 8-GPU node was available at submission time. The $250{,}000$-iteration run takes about one node-week of wall-clock time. Inference and evaluation use a single GPU.

\paragraph{Hyperparameters and search.}
We did not perform a hyperparameter search. All values listed in \cref{sec:training-setup} are taken from the NanoChat defaults for transformers of comparable depth; the only project-specific choices are the maximum sequence length ($4096$), the global batch size ($1{,}966{,}080$ tokens), and the number of iterations ($250{,}000$). The trace generator's own knobs (line cap, plan-operation range, noise mixture) are listed in \cref{sec:data-generation}.

\paragraph{Training-data volume.}
Training data is procedurally generated on the fly by the two samplers of \cref{sec:data-generation}, so no fixed training file is stored. Over $250{,}000$ optimization steps the model observes on the order of $5\times 10^{11}$ training tokens drawn from this distribution.

\paragraph{Validation and evaluation splits.}
The validation set used for monitoring during training is a disjoint seed range of the same generator and contains $1{,}024$ programs; it is regenerated deterministically from a fixed RNG state and is never used for model selection (we report the final-iteration checkpoint). The held-out evaluation suite consists of $252$ author-written MicroPy programs pooled across six tasks (\cref{tab:bitlen-stats}, \cref{tab:sat-stats}); none of these programs appears in the generator's support up to a probability lower bound of $10^{-6}$.

\paragraph{Statistics reporting.}
The headline accuracy numbers report token-level greedy-decoding accuracy on a single training seed; we do not aggregate across seeds because all reported numbers are $100\%$ and have zero variance under deterministic decoding. \Cref{fig:pencil-space-efficiency} reports per-task max trace lengths and is computed over the full $252$-program suite.

\paragraph{Software.}
We use PyTorch with the NanoChat training stack of \citet{karpathy2025nanochat} (MIT-licensed open source), which in turn uses the Muon optimizer of \citet{jordan2024muon}. The PENCIL scaffolding rule is implemented in our trace generator following \citet{yang2025pencil}. Tokenization uses a custom fixed-vocabulary whitespace tokenizer (described in \cref{sec:training-setup}). All third-party packages are used consistent with their published license and intended use.

\clearpage
\section{Extended Related Work}
\label{app:related-work}
\paragraph{Transformers, intermediate generation, and bounded context.}
Transformers augmented with intermediate generation can in principle carry out inherently serial computation, but the generated scratchpad itself becomes a memory bottleneck because it must remain accessible in context \citep{Vaswani2017attention,wei2022chain,merrill2023expresssive,li2024chain,liu2024lost}.
PENCIL is the closest scaffold to our setting: it reclaims completed reasoning steps so longer computations can proceed under a bounded context budget \citep{yang2025pencil}. MicroPy builds on this idea, using PENCIL to keep only the execution state needed for the next small-step rewrite rather than an ever-growing free-form scratchpad.

\paragraph{Length generalization by decomposition.}
A central difficulty in algorithmic reasoning is length generalization: models that fit short traces or short arithmetic instances often fail to extrapolate systematically to longer ones \citep{zhou2023algorithms,zhou2024robust}. Structural recursion is another persistent failure mode without explicit stack discipline \citep{zhang2024transformer}. Recent work shows that progress is possible when the task is reformulated so that each predicted step is more local: inductive scratchpads reduce global dependencies \citep{abbe2024globality}, while arithmetic-specific positional schemes such as position coupling and abacus-style embeddings improve extrapolation by aligning digit structure \citep{cho2024positioncoupling,mcleish2024rightembeddings}. MicroPy shares this decomposition principle, but applies it at the level of a programming language: instead of designing a better representation for one task such as addition, we train a general interpreter whose primitive predictions are reusable retrieval and rewrite steps.

Turing Programs is especially close in spirit: it shows that transformers can achieve robust length generalization when an algorithm is decomposed into Turing-machine-style updates that mostly copy the previous state with small local changes \citep{hou2024turingprograms}. The common principle is that out-of-distribution computations become sequences of simple in-distribution steps. The difference is that Turing Programs trains one model per task, whereas MicroPy trains one model to read prompt-defined procedures and execute unseen programs compositionally under a fixed language semantics.

\paragraph{Program-like analyses of transformers.} A complementary line writes, compiles, or recovers computations in transformer-oriented program form. RASP-style analyses study how transformer computations can implement symbolic algorithms, Tracr compiles such programs into transformer weights, and the Learning Transformer Programs work learns modified transformers that can be converted into discrete human-readable programs, including algorithmic tasks \citep{weiss2021thinking,lindner2024tracr,friedman2024learning}. These works make useful comparisons because they expose program structure inside or as task-specific transformer solutions, but they differ from MicroPy in the direction of construction: they compile, analyze, or recover transformer programs, whereas MicroPy trains a standard decoder-only transformer from random MicroPy execution traces and asks whether it learns reusable small-step interpreter rules for prompt-defined procedures under a fixed operational semantics.

\paragraph{Neural Programmer-Interpreters and learned execution.}
Neural Programmer-Interpreters (NPI) introduced learned execution with a recurrent controller, a learned program memory, and domain-specific interfaces, and later work showed gains from recursion and search in this framework \citep{npi_reed2016programmer_interpreters,npi_cai2017recursion,npi_pierrot2019alphanpi}. Architectural precursors with explicit external memory \citep{graves2014neural,graves2016hybrid} likewise targeted learnable universal computation, but routed control through differentiable memory rather than a discrete trace. Another line trains models more directly on execution traces or intermediate computations, from simple learned execution tasks to scratchpad supervision and Python trace prediction \citep{npi_zaremba2014learning_execute,nye2021show,npi_liu2023codeexecutor}. Our setting differs in that the model reads explicit MicroPy definitions in context, represents bindings, mutable state, and stack frames directly in the prompt, and is supervised on small-step retrieval and rewrite transitions compressed with PENCIL. The goal is not unrestricted program understanding, but a controlled test of whether a standard transformer can learn reusable interpreter rules for a small language.

\paragraph{Universal prediction from latent programs.}
\citet{graumoya2024learning} study a different use of universal computation: BrainPhoque universal-machine programs generate output sequences, and sequence models are meta-trained to predict the next emitted symbol as a finite approximation to universal prediction. The source program and internal machine configuration remain latent to their predictor. In MicroPy, the procedure definitions, bindings, mutable state, and active execution frames are explicit model inputs, and the target is the next interpreter transition rather than the next symbol emitted by an unknown program. PENCIL then repeatedly invokes the model and deterministically reclaims completed context; it does not choose the MicroPy semantic transition.

\paragraph{Algorithmic reasoning and in-context-learned algorithms.}
Several lines of work train neural models to execute structured procedures step by step. Symbolic-mathematics datasets train sequence models to perform multi-step arithmetic or symbolic computation \citep{saxton2019analysing,lample2020deep}. The Neural Algorithmic Reasoning programme trains a generalist model to imitate a suite of classical algorithms in the CLRS-30 benchmark \citep{velickovic2021neural,ibarz2022generalist}. A complementary thread reverse-engineers what algorithms are being learned in context \citep{akyurek2023learning,garg2022transformers,vonoswald2023transformers}. Architectural work on adaptive control flow inside the transformer \citep{csordas2022neural} addresses similar systematic-generalization concerns. MicroPy differs from this line on two axes: (i) we train one transformer on random execution traces of a Turing-complete language and evaluate on held-out human-written \emph{programs} rather than held-out instances of fixed algorithms, and (ii) we supervise on small-step retrieval/rewrite transitions rather than on input-output pairs. A complementary recent direction is to internalise procedural rules into the model weights from declarative instructions rather than execution examples \citep{cook2025programming}; that approach trades context-trace verbosity for finetuning-time compute, while MicroPy keeps the entire interpreter behaviour in-context and in-weights jointly.

\paragraph{Compositional generalization.}
MicroPy is also a test of compositional generalization: can a model execute novel combinations of familiar primitives and control-flow patterns rather than merely extend in-distribution traces? This theme is classically studied with SCAN, CFQ, and COGS-style splits that hold out compositions while keeping atoms familiar \citep{lake2018generalization,keysers2020measuring,kim2020cogs}. Our setting differs in that the composition rule is an operational semantics: generalization means recombining learned evaluation rules into correct executions on unseen programs, intersecting with broader limits of autoregressive transformers on multi-step compositional problems \citep{drozdov2022compositional,dziri2023faith}.

\paragraph{Retrieval augmentation, memory, and context.}
External-memory approaches such as retrieval augmentation and long-term memory modules address bounded context by storing information outside the immediate prompt \citep{lewis2020rag,wang2024augmenting}. These approaches are complementary to MicroPy: our focus is not how to retrieve external knowledge, but which execution state must stay in context so that the next interpreter step is well-defined.

\clearpage
\section{Example Function Walkthrough}
\label{app:example-data}
The training data generator is fairly complex.  Rather than giving a full specification for all the language features
we give an example of evaluation and leave it to the reader to imagine a generalization to the full language.
Consider two definitions of the identity function.

\begin{verbatim}
@MicroPy
def foo(x): return x

@MicroPy
def bar(x): return x
\end{verbatim}

We will generate training data by evaluating the expression {\tt foo(bar(obj1))}.  This generates a training context that includes
the definitions of {\tt foo} and {\tt bar} as

\begin{verbatim}
FD foo = lambda(x) Exp3;
FD bar = lambda(x) Exp4;
\end{verbatim}

The training sequence also includes the following expression definitions.

\begin{verbatim}
D.Exp1 = App(bar (O.obj1))
D.Exp2 = App(foo (Exp1))
D.Exp3 = LookupVar(x)
D.Exp4 = LookupVar(x)
\end{verbatim}

The first training sequence starts with a call to an expression

\begin{verbatim}
[call] [call] Exp2
\end{verbatim}

As will become clear, the two occurrences of {\tt [call]} are used to combine effects.  In this example there are no
effects but the evaluation rules handle the general case.

This initial context is followed by token generation resulting
in a PENCIL rewrite of {\tt Exp2} to its definition.
\footnote{The examples of token sequences are formatted here to improve reading.  In the actual context there are no newlines or tabs.}
\begin{verbatim}
[call] [call] Exp2
  => [call] App(foo (Exp1)) [ret]
\end{verbatim}

PENCIL rewriting then gives:

\begin{verbatim}
[call] [call] App(foo (Exp1))
\end{verbatim}

Continuing with token generation we get.

\begin{verbatim}
[call] [call] App(foo (Exp1))
 arg0= [call] [call] Exp1
\end{verbatim}

Token generation then yields a rewriting of {\tt Exp1} to its definition.

\begin{verbatim}
[call] [call] App(foo (Exp1))
 arg0= [call] [call] Exp1
  => [call] App(bar (O.obj1)) [ret]
\end{verbatim}

Which yields:

\begin{verbatim}
[call] [call] App(foo (Exp1))
 arg0= [call] [call] App(bar (O.obj1))
\end{verbatim}

Continued token generation yields:

\begin{verbatim}
[call] [call] App(foo (Exp1))
 arg0= [call] [call] App(bar (O.obj1))
  arg0 = [call] O.obj1
\end{verbatim}

Only single {\tt [call]} occurs when evaluating an object constant (no effect is involved).
The innermost call is then replaced by an effect and a value.

\begin{verbatim}
[call] [call] App(foo (Exp1))
 arg0= [call] [call] App(bar (O.obj1))
  arg0 = Eff(empty) O.obj1
\end{verbatim}

The next step is to retrieve the definition of the procedure bar and replace the call to bar
with a stack frame in which to evaluate the body expression. This is done in three steps: the procedure definition is retrieved; the effect of the argument evaluation is
separated from the evaluation of the procedure body; and a stack frame is constructed for the evaluation of the body.
Here we show the context after the construction of the stack frame.

\begin{verbatim}
[call] [call] App(foo (Exp1))
  arg0= [call] Eff(empty)
    [call] Env(Bind(x O.obj1))
      [call] Exp4
\end{verbatim}

From here {\tt Exp4} is rewritten to its definition.

\begin{verbatim}
[call] [call] App(foo (Exp1))
  arg0= [call] Eff(empty)
    [call] Env(Bind(x O.obj1))
      [call] LookupVar(x)
\end{verbatim}

Then the Lookup is evaluated.

\begin{verbatim}
[call] [call] App(foo (Exp1))
  arg0= [call] Eff(empty)
    [call] Env(Bind(x O.obj1))
      Eff(empty) O.obj1
\end{verbatim}

Next, the innermost call (the stack frame) is replaced by its effect and value.

\begin{verbatim}
[call] [call] App(foo (Exp1))
  arg0= [call] Eff(empty)
               Eff(empty)
               O.obj1
\end{verbatim}

Next the two effects are combined.
\begin{verbatim}
[call] [call] App(foo (Exp1))
  arg0= Eff(empty) O.obj1
\end{verbatim}

From here the computation proceeds similarly to the computation of the application of {\tt bar}.

\clearpage
\section{MicroPy Function Definitions}
\label{app:micropy-functions}
{\scriptsize
\begin{Verbatim}[breaklines=true,breakanywhere=true]
@MicroPy
def copy_bits(b1,b2):
    return Seq(Assert(b2,Attr("value"),LookupAttr(b1,Attr("value"))),
                If(HasAttr(b1,Attr("next")),
                    copy_bits(LookupAttr(b1,Attr("next")),
                              LookupAttr(b2,Attr("next"))),
                    unit_))

@MicroPy
def not_(p):
    return If(p,false_,true_)

@MicroPy
def flip_bits(b1,b2):
    return Seq(Assert(b2,Attr("value"),not_(LookupAttr(b1,Attr("value")))),
                If(HasAttr(b1,Attr("next")),
                    flip_bits(LookupAttr(b1,Attr("next")),
                                LookupAttr(b2,Attr("next"))),
                    unit_))

@MicroPy
def exor_(p,q):
    return If(p,If(q,false_,true_),If(q,true_,false_))

@MicroPy
def and_(p,q):
    return If(p,q,false_)

@MicroPy
def or_(p,q):
    return If(p,true_,q)

@MicroPy
def rpc_sum_(a1,a2,c):
    return exor_(a1,exor_(a2,c))

@MicroPy
def rpc_carry_(a1,a2,c):
    return or_(and_(a1,a2),
               and_(c,or_(a1,a2)))

@MicroPy
def RPC_add(b1,b2,b3,c):
    return RPC_aux(b1,b2,b3,c,
                   rpc_carry_(LookupAttr(b1,Attr("value")),
                              LookupAttr(b2,Attr("value")),
                              c))

@MicroPy
def RPC_aux(b1,b2,b3,c,nc):
    return Seq(Assert(b3,Attr("value"),
                      rpc_sum_(LookupAttr(b1,Attr("value")),
                               LookupAttr(b2,Attr("value")),
                               c)),
                If(HasAttr(b1,Attr("next")),
                    RPC_add(LookupAttr(b1,Attr("next")),
                            LookupAttr(b2,Attr("next")),
                            LookupAttr(b3,Attr("next")),
                            nc),
                    Assert(LookupAttr(b3,Attr("next")),
                           Attr("value"),
                           nc)))

@MicroPy
def RPC_mult(b1,b2,b3):
    return Seq(If(LookupAttr(b1,Attr("value")),
                    RPC_add(b2,b3,b3,false_),
                    unit_),
                If(HasAttr(b1,Attr("next")),
                    RPC_mult(LookupAttr(b1,Attr("next")),
                             b2,
                             LookupAttr(b3,Attr("next"))),
                    unit_))

@MicroPy
def sat_assign(var, cnf_head_, out_):
    return If(
        HasAttr(var, Attr("next")),
        Try(
            Seq(Assert(var, Attr("value"), true_),
                sat_assign(LookupAttr(var, Attr("next")),
                           cnf_head_,
                           out_)),
            Seq(Assert(var, Attr("value"), false_),
                sat_assign(LookupAttr(var, Attr("next")),
                           cnf_head_,
                           out_)),
        ),
        Try(
            Seq(
                Assert(var, Attr("value"), true_),
                If(eval_cnf(cnf_head_),
                   Assert(out_, Attr("value"), true_),
                   fail_),
            ),
            Seq(
                Assert(var, Attr("value"), false_),
                If(eval_cnf(cnf_head_),
                   Assert(out_, Attr("value"), true_),
                   fail_),
            ),
        ),
    )

@MicroPy
def sat_solve(var_head_, cnf_head_, out_):
    return Try(sat_assign(var_head_, cnf_head_, out_),
               Assert(out_, Attr("value"), false_))
\end{Verbatim}
}

\clearpage
\section{Frontier-LLM Probe on MicroPy Traces}
\label{app:llm-probe}

We gave Claude Opus~4.7 read access to the MicroPy source code and asked it to predict step-by-step completions on a balanced sample of $996$ held-out trace steps ($166$ per task). The model was free to use tools, and in this setup chose to do so: it read the operational semantics from the source, wrote its own Python re-implementation of the MicroPy rewrite rules, and invoked that re-implementation as a tool to produce predictions. The probe therefore measures what a tool-using frontier LLM achieves when it tries to \emph{simulate} MicroPy's trace generator from the source documentation, rather than what an LLM does by next-token reasoning alone. Gold completions are withheld from the model's input.

\begin{table}[h]
\centering
\small
\setlength{\tabcolsep}{8pt}
\begin{tabular}{@{}lrr@{}}
\toprule
Task & $n$ & Exact-match \\
\midrule
RPC\_add        & $166$ & $0.777$ \\
RPC\_mult       & $166$ & $0.747$ \\
flip\_bits      & $166$ & $0.729$ \\
sat\_solve      & $166$ & $0.741$ \\
sat\_verify     & $166$ & $0.693$ \\
copy\_bits      & $166$ & $0.669$ \\
\midrule
\textbf{Overall} & $\mathbf{996}$ & $\mathbf{0.726}$ \\
\bottomrule
\end{tabular}
\caption{Claude Opus~4.7 with source-code access and self-authored tool use on MicroPy step-prediction ($n=996$, balanced across the six held-out tasks). The remaining error stems from places where the model's hand-written Python simulator diverges from MicroPy's actual semantics, particularly on accumulated in-trace effects.}
\label{tab:llm-probe}
\end{table}

We read this number two ways. First, MicroPy step-prediction is not a string-pattern task a pretrained LLM incidentally solves: even with full language documentation and a self-written Python re-implementation invoked as a tool, a frontier LLM does not match the trained interpreter's $100\%$. Second, this is \emph{not} a task-performance comparison. A tool-using LLM agent could trivially answer the underlying bit-arithmetic and SAT questions by invoking Python or Z3, and (as observed here) it can even author the Python re-implementation used in this probe. The relevant comparison is therefore not "LLM with tools" against our trained interpreter but "LLM \emph{without} external solvers, forced to perform extended arithmetic or SAT directly"; existing length-generalization work \citep{zhou2023algorithms,zhou2024robust,seely2025sudokubench,hazra2024threesat} reports that pretrained LLMs do not generalize in this regime. Our contribution is specifically about whether a transformer can be trained to internalize a Turing-complete operational semantics from execution traces, which is orthogonal to whether a different system, given different affordances, could solve the underlying tasks.

\clearpage
\section{Distribution-Shape Failure Examples}
\label{app:failure-case-tail-eff}

As MicroPy designers, we anticipated two axes along which a held-out test program can exercise a structural shape that a sparser training distribution would miss: (i) long single-step writes inside an effect-accumulation step, and (ii) zero-argument call boundaries. The training generator used for the model reported in \cref{sec:results} explicitly samples both axes. To illustrate \emph{what an OOD failure on these axes looks like}, we describe below two examples produced by an ablated training distribution that under-samples each axis. These examples are not the model that produces the headline numbers; they are case studies of the failure mode an under-covered distribution induces.

\subsection*{Failure 1: Long Pre-\texttt{TailApp} \texttt{Eff} write exceeding training write length}

\paragraph{Setup.}
We constructed a held-out program whose execution requires a tail-call \texttt{[ret]} step that closes two adjacent \texttt{Eff} frames into a single combined $\sim$$52$-entry \texttt{Eff} immediately preceding a \texttt{TailApp}, after extended linked-list traversal accumulates many \texttt{Assertion(o,\,attr0,\,bool)} writes. At the token level this is a long copy-with-merge: the model must reproduce dozens of \texttt{Assertion(\ldots)} entries in order, drawn from the two upstream \texttt{Eff} frames in the prompt tail, then re-emit the \texttt{TailApp} payload unchanged.

The write length here exceeds the longest single-step write the model saw during training. The immediate prompt tail is:

\begin{small}
\begin{verbatim}
general stack (prompt tail):
  [call] Env ()
  [call] Eff ( Assertion(obj2  attr0 False)
               Assertion(obj51 attr0 False)
               ... 37 more Assertions ...
               Assertion(obj34 attr0 False) )
         Eff ( Assertion(obj45 attr0 False)
               Assertion(obj44 attr0 True)
               ... 11 more Assertions ...
               Assertion(obj33 attr0 True) )
         TailApp(fun7 Args(obj6 obj1 obj44))   [EOP]
\end{verbatim}
\end{small}

\paragraph{Gold vs.\ prediction.}
The gold completion merges the two upstream \texttt{Eff} frames into one combined $52$-entry \texttt{Eff} and forwards the \texttt{TailApp} payload to the parent step:

\begin{small}
\begin{verbatim}
Gold:  => [call] Eff (
            Assertion(obj2  attr0 False)
            ... position 108: obj47 attr0 True ...
            ... 51 more Assertions ...
            Assertion(obj33 attr0 True) )
          TailApp(fun7 Args(obj6 obj1 obj44)) [ret]

Model: => [call] Eff (
            Assertion(obj2  attr0 False)
            ... position 108: obj182 attr0 True ...
            ... 51 more Assertions ...
            Assertion(obj33 attr0 True) )
          TailApp(fun7 Args(obj6 obj1 obj44)) [ret]
\end{verbatim}
\end{small}

The first 107 tokens match exactly; at token position $108$ (inside the $18^{\text{th}}$ \texttt{Assertion} slot of the combined \texttt{Eff}) the gold token is \texttt{obj47} but the model emits \texttt{obj182}. The remaining $\sim$$200$ tokens of the long completion are regenerated correctly. The error is silent: structurally the output is a well-formed \texttt{Eff(\ldots) TailApp(\ldots) [ret]} step, but one object reference inside the accumulator has been swapped to a different valid-looking object name. \texttt{obj47} appears explicitly in the immediate prompt tail (the gold token is a direct copy), yet \texttt{obj47} is not in the top-$10$ predictions at this position.

\paragraph{Interpretation.}
A 52-entry combined \texttt{Eff} write is $\sim$$300$ tokens of structural copy, supervised only by the end-to-end completion loss. When this length exceeds the longest combined-\texttt{Eff} write present in training data, per-slot error rate compounds across the length: the model gets the structural skeleton, attribute names, and bool literals right, but the obj-name slot, which has tens of equally-typed candidates in the immediate context, becomes weakly identified.

\paragraph{Curriculum coverage in the reported model.}
Because long pre-\texttt{TailApp} \texttt{Eff}-merge writes are a foreseeable failure axis given MicroPy's stack discipline, the training generator used for the model in \cref{sec:results} explicitly samples such long writes with distractor object names sharing the same attribute. On the same example the reported model produces the gold token \texttt{obj47} at position $108$.

\subsection*{Failure 2: Zero-argument function calls (dummy-arg coverage gap)}

\paragraph{Setup.}
Several natural held-out programs include helper procedures that, semantically, take no arguments. For example, a top-level entry such as
\begin{small}
\begin{verbatim}
@MicroPy
def helper():
    ...
\end{verbatim}
\end{small}
\noindent compiles into MicroPy traces whose call sites are \texttt{App(funX ())} and \texttt{TailApp(funX Args())}: function tokens with an immediately closing empty argument list.

The training distribution did not exercise the zero-argument shape: every procedure in the sampled programs had at least one variable bound at call time. The empty-argument-list token sequence (\texttt{Args ( )} adjacent in the trace) was therefore essentially out-of-distribution at the structural level.

\paragraph{Observed failure.}
On a held-out program that uses such a zero-argument helper, the original definition produced $7$ token-level misses across $3544$ trace rows at our reference checkpoint, all clustered around the \texttt{App(funX ())} / \texttt{TailApp(funX Args())} call boundaries. The whole-program success rate was $0/1$.

\paragraph{Curriculum coverage in the reported model.}
The training generator used for the model in \cref{sec:results} treats the zero-argument call boundary as a first-class shape and samples \texttt{App(funX ())} / \texttt{TailApp(funX Args())} explicitly. On the same program, the reported model produces no misses at the zero-argument call boundaries.

\paragraph{Combined takeaway.}
Both examples are consistent with the same pattern. When the trained model is presented with a structural shape its training distribution did not cover, the model degrades gracefully \emph{into the relevant prior} rather than failing structurally. The reported model in \cref{sec:results} covers both axes by design, but we include these examples to make the under-coverage failure mode concrete.

\clearpage
\begin{figure*}[!ht]
  \section{Per-Task Evaluation Operation Profiles}
  \label{app:per-task-operation-profiles}
  \centering
  \begin{subfigure}[b]{0.32\textwidth}
    \centering
    \includegraphics[width=\textwidth]{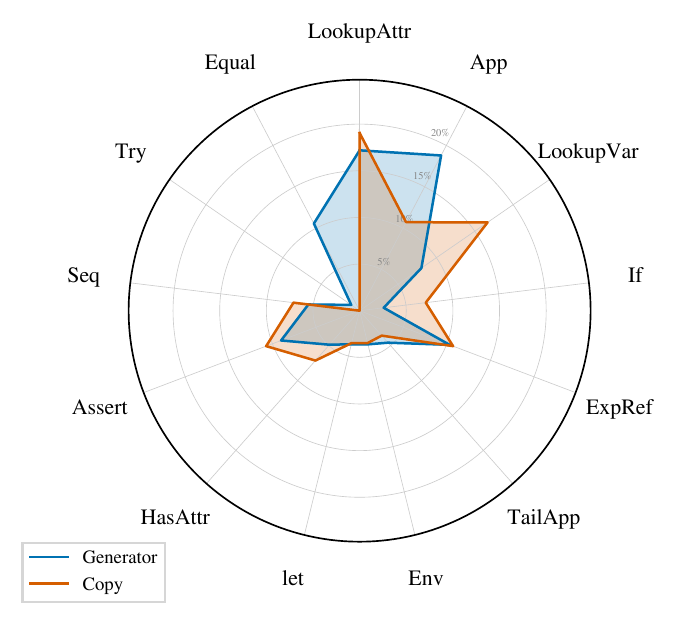}
    \caption{Copy}
  \end{subfigure}\hfill
  \begin{subfigure}[b]{0.32\textwidth}
    \centering
    \includegraphics[width=\textwidth]{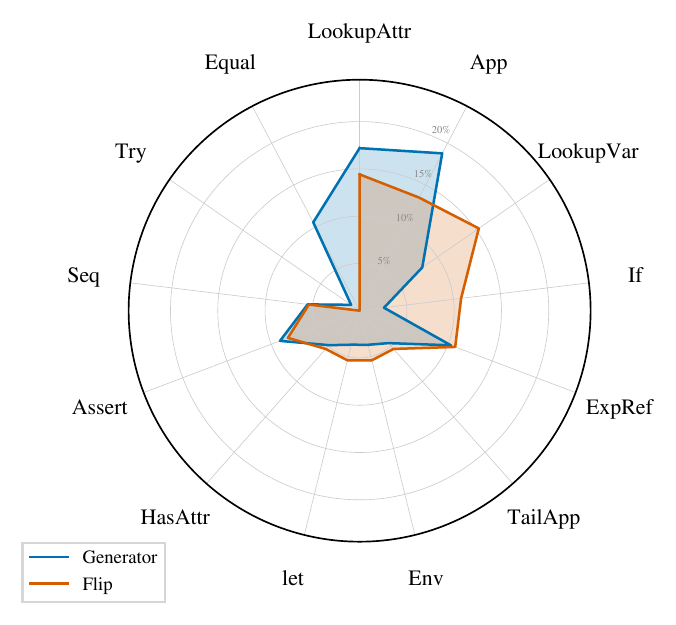}
    \caption{Flip}
  \end{subfigure}\hfill
  \begin{subfigure}[b]{0.32\textwidth}
    \centering
    \includegraphics[width=\textwidth]{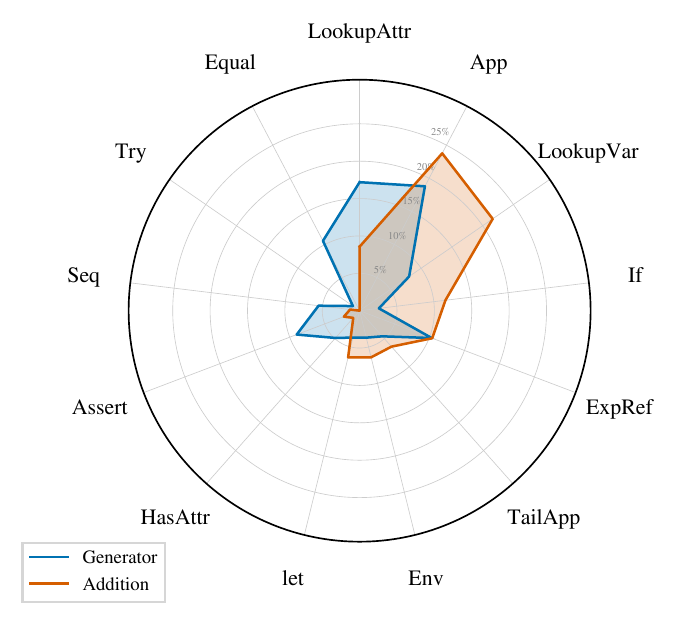}
    \caption{Addition}
  \end{subfigure}

  \begin{subfigure}[b]{0.32\textwidth}
    \centering
    \includegraphics[width=\textwidth]{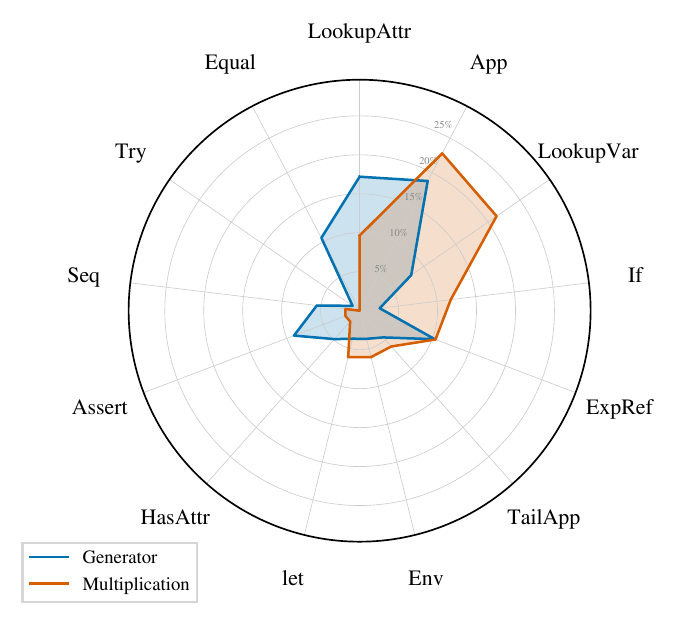}
    \caption{Multiplication}
  \end{subfigure}\hfill
  \begin{subfigure}[b]{0.32\textwidth}
    \centering
    \includegraphics[width=\textwidth]{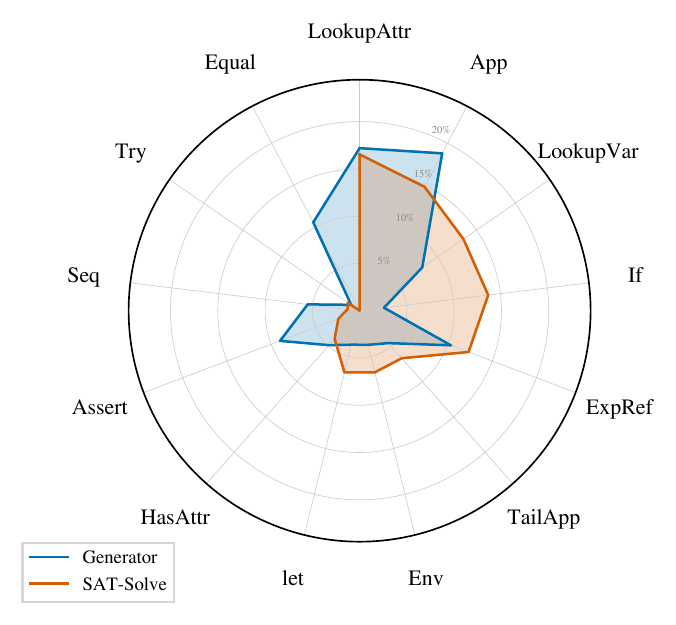}
    \caption{SAT-Solve}
  \end{subfigure}\hfill
  \begin{subfigure}[b]{0.32\textwidth}
    \centering
    \includegraphics[width=\textwidth]{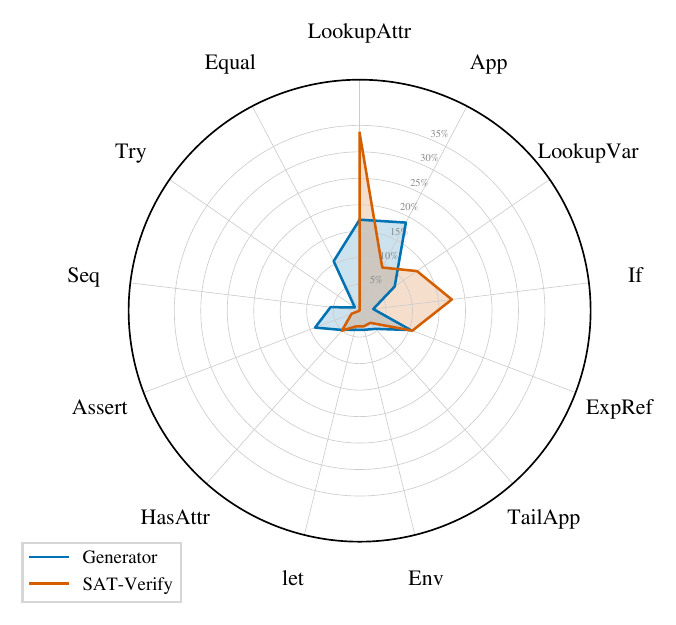}
    \caption{SAT-Verify}
  \end{subfigure}
  \caption{Per-task operation profiles for the held-out evaluation suite. Each subplot compares the training generator's last-op distribution (blue) against a single evaluation task (orange). Proportions are within-corpus. Copy and Flip are dominated by \texttt{App}/\texttt{LookupAttr}/\texttt{ExpRef} patterns from linked-list traversal; Addition and Multiplication rely more heavily on \texttt{If}/\texttt{LookupVar}; and the SAT tasks show the expected \texttt{Try}-heavy profile from backtracking search.}
  \label{fig:per-task-radar}
\end{figure*}

\end{document}